\documentclass[journal]{IEEEtran}
\usepackage{graphicx}
\usepackage{subfigure}
\usepackage{multirow}
\usepackage{longtable}
\usepackage{algorithm}
\usepackage{algorithmic}
\usepackage{booktabs}
\usepackage{stfloats}
\usepackage{caption}
\usepackage{amssymb,amsmath}
\usepackage{color}
\usepackage{fancyheadings}
\usepackage{hyperref}
\usepackage{cite}

\begin{document}
\title{Rethinking Gradient Operator for Exposing AI-enabled Face Forgeries}

\author{Zhiqing~Guo$^{1}$  \quad Gaobo~Yang$^{1}$ \quad Dengyong~Zhang$^{2}$ \quad Ming~Xia$^{1}$ \qquad \vspace{1pt}\\
$^{1}$Hunan University  \qquad $^{2}$Nanjing University of Information Science and Technology\qquad\qquad\\
\hspace{0.1in}{\tt\small \{guozhiqing, yanggaobo\}@hnu.edu.cn} \qquad {\tt\small zhdy@csust.edu.cn} \qquad {\tt\small mingxia@163.com} \\
}

\markboth{}%
{Shell \MakeLowercase{\textit{et al.}}: Rethinking Gradient Operator for Exposing AI-enabled Face Forgeries}

\maketitle

\renewcommand{\headrulewidth}{0pt}

\begin{abstract}
For image forensics, convolutional neural networks (CNNs) tend to learn content features rather than subtle manipulation traces, which limits forensic performance. Existing methods predominantly solve the above challenges by following a general pipeline, that is, subtracting the original pixel value from the predicted pixel value to make CNNs pay attention to the manipulation traces. However, due to the complicated learning mechanism, these methods may bring some unnecessary performance losses. In this work, we rethink the advantages of gradient operator in exposing face forgery, and design two plug-and-play modules by combining gradient operator with CNNs, namely tensor pre-processing (TP) and manipulation trace attention (MTA) module. Specifically, TP module refines the feature tensor of each channel in the network by gradient operator to highlight the manipulation traces and improve the feature representation. Moreover, MTA module considers two dimensions, namely channel and manipulation traces, to force the network to learn the distribution of manipulation traces. These two modules can be seamlessly integrated into CNNs for end-to-end training. Experiments show that the proposed network achieves better results than prior works on five public datasets. Especially, TP module greatly improves the accuracy by at least 4.60\% compared with the existing pre-processing module only via simple tensor refinement. The code is available at: https://github.com/EricGzq/GocNet-pytorch.
\end{abstract}
\begin{IEEEkeywords}
Attention mechanism, deepfake, face forgery detection, gradient operator.
\end{IEEEkeywords}
\IEEEpeerreviewmaketitle

\section{INTRODUCTION}
\IEEEPARstart{p}{eople} have an intuitive and pre-conceived belief that seeing is believing. However, with the proliferation of face image forgery techniques, fake face images have brought serious social problems. For example, an easy-to-use AI-based App, which can generate photo-realistic face images from the faces of celebrities and politicians, might be used to mislead social opinions\footnote{https://www.theregister.co.uk/2018/01/25/ai\_fake\_skin\_flicks/}. Thus, there is an urgent need to develop  some detectors to expose face forgeries.

In recent years, face image forgery detection has attracted wide attentions in the community of image forensics. Different from the hand-crafted features, the convolutional neural networks (CNNs) based works learn features in an end-to-end manner to expose face forgeries \cite{MesoNet}. However, CNNs tend to learn image content features rather than subtle manipulation traces, which constrains both detection accuracy and robustness \cite{constrainedCNN}. Some works have attempted to exploit pre-processing to suppress image content and highlight manipulation traces, which improves detection performance. The existing pre-processing methods for the CNN-based image forensics can be divided into two categories: a) pre-processing image data in an off-line way and inputting the pre-processed data into the deep model for feature learning \cite{chen2019attention,HPcnn};
b) adding some constrains on the first convolution layer to enforce the CNNs to learn features from manipulation traces \cite{constrainedCNN,AMTEN}.
In fact, the former highlights the manipulation traces, which can be well observed from the pre-processed data. Thus, deep models can naturally learn the representation of manipulation traces. This method has good interpretability, but the off-line way is quite time-consuming. The latter simplifies network training by keeping the end-to-end way, but we not sure whether manipulation traces have been properly captured or not by CNNs due to the black box nature. An intuitive yet important question is that can the classical image pre-processing methods be embedded into CNNs for efficient training while keeping good interpretability?

\begin{figure*}
  \centering
  \includegraphics[width=6.5in]{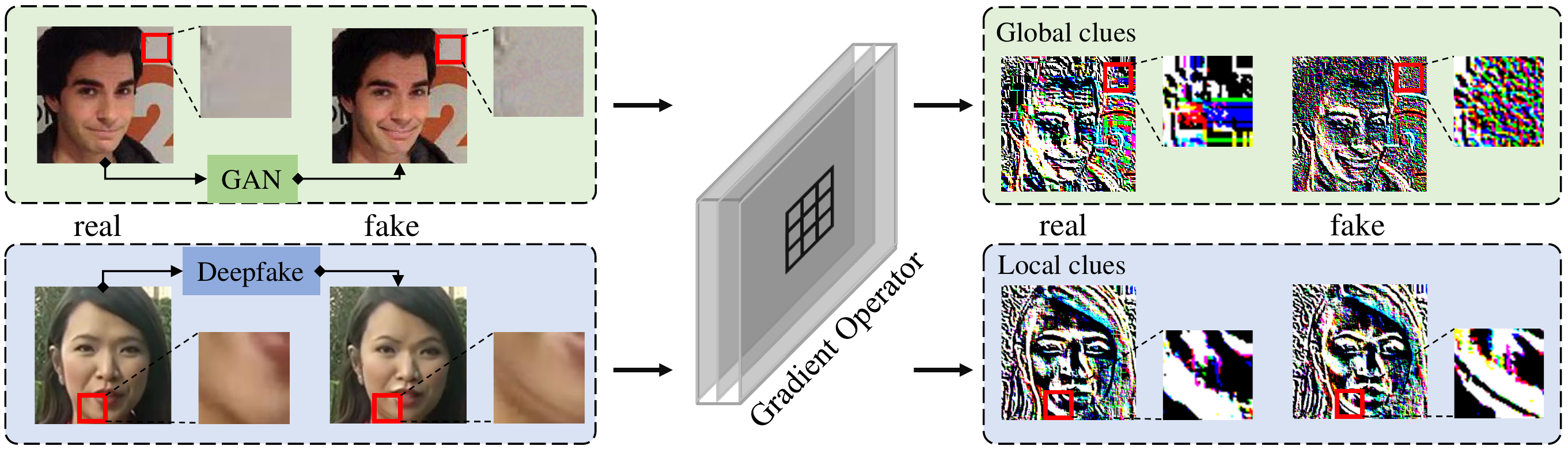}
  \caption{Using gradient operator highlights the manipulation clues left by face forgeries. Note that global clues are generally widely distributed in global images, while local clues are only distributed in the swapped faces and blending boundaries.}\label{show_traces}
\end{figure*}

Some recent studies have claimed that generative adversarial networks (GANs) leave inherent fingerprints in GAN-generated global images \cite{Fingerprints,DoGANs}. In addition, some face-swapping techniques such as DeepFake leave local artifacts near blending boundaries \cite{FaceXray}. Either global fingerprints or local artifacts are key manipulation traces to expose face forgeries. In the field of image processing, the gradient operators are widely-used for locating image edges with gray-scale changes. Actually, manipulation traces can also be effectively captured by the gradient operators. As shown in Fig. \ref{show_traces}, it is difficult to observe the subtle differences between real and fake face images by naked eyes, but the invisible manipulation traces are effectively highlighted after pre-processing with the gradient operator.
Nevertheless, Mo et al. \cite{HPcnn} have used the gradient operators such as high-pass filter to process candidate face images, but without considering the efficient combination with CNNs. In addition, Bayar et al. \cite{constrainedCNN} highlighted manipulation traces by changing some kernel coefficients in the convolution layer. However, most kernel coefficients still need to be adaptively learned via back propagation, which brings uncertainty to the extracted manipulation traces.
To design an efficient deep network with good interpretability, we propose to embed the classical gradient operator as the convolution layer with fixed kernel coefficients, which refines the tensors of different channels in the network model. Thus, a simple yet efficient pre-processing module is designed for the network to capture manipulation traces in an end-to-end manner.

Besides designing specific pre-processing for face forensics, some existing CNN-based detectors also force the network to pay attention to the local artifacts near blending boundary that are generated by specific face forgery techniques \cite{FaceXray}. However, these methods that only capture local artifacts still have some limitations, which make it impossible for the network to detect multiple forgery techniques simultaneously. Actually, no matter global fingerprints left by GANs or local artifacts near blending boundary are key manipulation traces, which should be highlighted so that the CNNs can effectively learn features from them for image forensics. We argue that capturing manipulation traces via global attention module is an effective way to expose various face forgeries. As we know, global attention modules have been widely used in computer vision tasks such as scene segmentation and image classification \cite{CBAM,DAnet}. However, the existing global attention mechanisms consider only from spatial and channel dimensions when improving feature representation. When designing the global attention module for face image forensics, we should consider the manipulation traces. Due to the superior performance of gradient operators in capturing manipulation traces, we design an attention module equipped with the classical gradient operators to capture global manipulation clues.

In this work, we distill the above insights and design two feature refinement modules, namely tensor pre-processing (TP) module and manipulation trace attention (MTA) module. Then, a gradient operator convolutional network (GocNet) is constructed by equipping both TP and MTA modules in a dual-stream network. Specifically, the main contributions of this work are three-fold.
\begin{itemize}
\item The TP module is designed for pre-processing, which embeds the classical gradient operator into the convolution network as a fixed convolution layer. This is an end-to-end learning method, which forces the network to learn features from key manipulation traces while ignoring irrelevant image content. Compared with the existing pre-processing methods, our method has no the inconvenience of off-line pre-processing, which means that it is more direct and effective.
\item The MTA module is designed for feature refinement, which combines the classical gradient operator with channel attention to capture the manipulation clues from global images. Different from the attention mechanisms in the existing face forensics works, the MTA module pays attention to the manipulation traces and artifacts in the global scope, instead of just local manipulation regions.
\item We propose a dual-stream GocNet, which is equipped with both TP and MTA modules, to expose various face image forgeries. It can achieve robust detection on five public fake face datasets. Extensive experimental results show that the proposed GocNet achieves much better performances than the state-of-the-art works.
\end{itemize}

\indent The remainder of this paper is organized as follows. Section \uppercase\expandafter{\romannumeral2} briefly introduces the related works. Section \uppercase\expandafter{\romannumeral3} presents the proposed TP module, MTA module and network structure of GocNet. Section \uppercase\expandafter{\romannumeral4} reports the experimental results with some analysis. Conclusion is made in Section \uppercase\expandafter{\romannumeral5}.

\section{Related Works}
\subsection{Face Forensics Works}
With the increasing concern about face image forgeries, there is a surge of interest in developing fake face detection methods.
Afchar et al. \cite{MesoNet} built two shallow networks via a few convolution layers, which exposed some early face forgery techniques. Yang et al. \cite{head_poses} exposed the Deepfake by using the inconsistency of 3D head posture. Li et al. \cite{SimulateArtifacts} exposed deepfake videos via the CNN by detecting face warping artifacts near the blending boundary. Matern et al. \cite{visual_artifacts} exploited simple visual artifacts to effectively expose the face forgeries such as DeepFake and Face2Face. However, these works expose fake face images only by either simply constructing the convolution networks or using the defects of biological features, without designing specific modules to improve the feature representation of the networks by capturing manipulation traces.

\subsection{Pre-processing Module}
To effectively learn features from manipulation traces, some works have introduced the pre-processing methods into the CNNs. For example, Chen et al. \cite{chen2019attention} proposed to convert RGB images into multi-scale retinex images, which captures well high-frequency information, to expose face spoofing detection. Mo et al. \cite{HPcnn} exploited the high pass filter to highlight the manipulation traces, so as to expose GAN-generated fake face images.
However, these two works process data in an off-line way, which is very time-consuming. Thus, some end-to-end methods were proposed to simplify the training process. For example, Bayar et al. \cite{constrainedCNN} developed a constrained convolution layer (C-Layer), which can suppress image content and highlight manipulation traces, for general purpose manipulation detection. Inspired by the existing image forensic methods, Guo et al. \cite{AMTEN} also designed an adaptive manipulation traces extraction network (AMTEN) to expose various face forgeries. These existing methods follow similar idea that obtaining residuals from prediction and learning features from the residuals for forensics. However, due to the diversity of face forgeries, it is difficult to accurately model and predict the manipulation traces, which inevitably leads to biased prediction.

\subsection{Feature Attention Module}
To further improve performance, some works embed the attention mechanism into the CNNs to enforce the network learn  discriminant features. For example, Woo et al. \cite{CBAM} designed a convolutional block attention module, which can be integrated into any CNNs to improve feature representation. Fu et al. \cite{DAnet} proposed a dual attention network for scene segmentation, in which two types of attention modules model the semantic interdependencies in spatial and channel dimensions, respectively. In the field of face forensics, Dang et al. \cite{dang2020detection} used the attention mechanism to visualize the manipulated regions and improve the feature maps for face forgery dection. Li et al. \cite{FaceXray} proposed an image representation called face X-ray for more general face forgery detection, in which local manipulation artifacts are learned via CNNs. As we know, various face forgeries leave manipulation traces distributing either globally or locally. If the attention module focuses on only local manipulation clues, it will make the network unable to detect various face forgeries.

\section{Methodology}
As analyzed in the section of related works, the prediction-based residuals suffer from possible bias, which has side effects on successive feature learning. Moreover, the attention mechanism should pay attention to global and local artifacts, so as to expose various face forgeries. Instead of using the prediction pipeline, we propose to embed the classical gradient operator, which serves as the convolution layer with fixed kernel coefficients, into the deep network. That is, the gradient operator is used to refine the input feature tensor, which is actually the TP module, to highlight manipulation traces. We also design an MTA module to capture global manipulation traces for exposing various face forgeries. In our approach, we designed two modules, which can make the network pay attention to the manipulation traces, to improve the feature representation of the network.

\subsection{Tensor Pre-processing Module}
\begin{figure}
  \centering
  \includegraphics[width=3.2in]{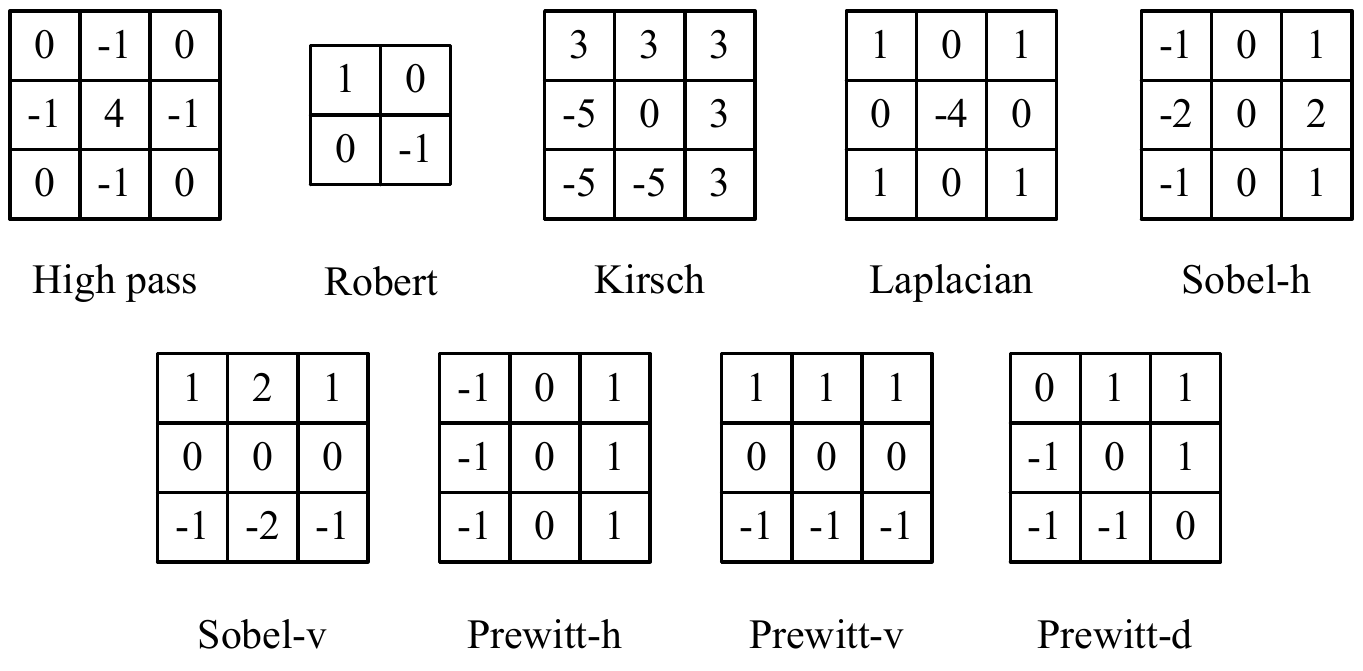}
  \caption{Nine classical gradient operators, in which `-h', `-v' and `-d' represent horizontal, vertical and diagonal directions, respectively.}\label{gradient_operators}
\end{figure}

For face forgeries, manipulation traces are usually presented as isolated dots or linear textures that are difficult for naked eyes to recognize. Thus, the manipulation traces are usually accompanied with the change of image gradients. Obviously, the classical gradient operators can be used to highlight the manipulation traces. In the network, the input image is transformed into the tensor $T_i$. Thus, the gradient operator $G$ is used to make convolution operation with the tensors of different channels. That is, the refined tensor $T_o$ is obtained as follows.
\begin{equation}\label{eq_1}
     T_{o}=\sum_{i=1}^{N}T_{i} \ast G
\end{equation}
where $N$ is the number of channels, $G$ denotes the possible gradient operators, as summarized in Fig. \ref{gradient_operators}. Unlike the adaptive learning methods such as the C-Layer \cite{constrainedCNN} and AMTEN \cite{AMTEN} for obtaining the kernel coefficients, the convolution operation in the TP module is equivalent to setting the convolution layer with fixed kernel coefficients. That is, the TP module highlights the manipulation traces with the help of the classical gradient operators, which is simple yet effective for pre-processing.

\subsection{Manipulation Trace Attention Module}
The existing attention mechanisms usually model the pixel interdependencies in spatial or channel dimensions to improve the representation of feature tensor. In the face forensics tasks, we also exploit the attention module to refine feature tensor, so that the network can learn more discriminative features. And the difference is that for face forensics tasks, more attention should be paid to manipulation traces. Thus, the design of the attention module should focus on manipulation traces, not just pixel interdependencies.

\begin{figure*}
  \centering
  \includegraphics[width=6.5in]{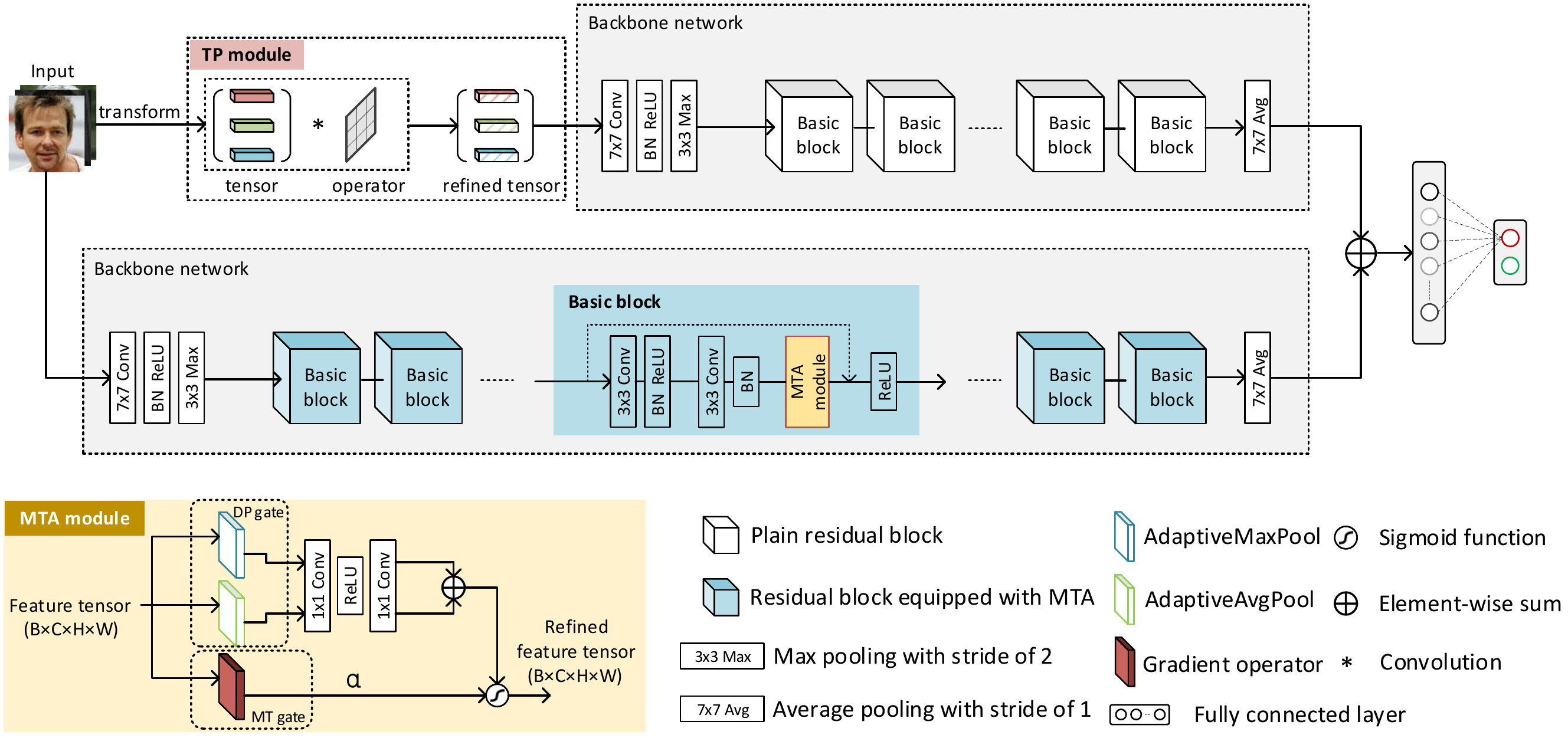}
  \caption{Overview of the proposed framework.}\label{architecture}
\end{figure*}

To capture well manipulation traces, our MTA module is designed from two dimensions, as shown in Fig. \ref{architecture}.
On the one hand, we argue that the manipulation traces between different channels are related. Thus, we construct a double pooling (DP) gate in the MTA module, following the double pooling operation in \cite{CBAM}. The DP gate captures the manipulation trace association between channels, and produces the channel attention map by the shared network.
On the other hand, we use the gradient operator to set the manipulation trace (MT) gate in the MTA module, which refines the feature map to remove irrelevant content information and highlight key manipulation traces.

Specifically, let $F \in \mathbb{R}^{C \times H \times W}$ be the given input feature map, which is fed into the MTA module to obtain a refined feature map $F'$. For the DP gate, an average pooling operation is used to learn the distribution range of the manipulation traces in the feature maps, and the max-pooling operation is used to learn unique trace features in different channels. Thus, the DP gate aggregates spatial information to infer finer channel-wise attention. Then, the pooled features are fed into the shared network $S_n$, which is composed of two $1\times1$ convolution layers and a ReLU activation function, to obtain the attention map $M_{c} \in \mathbb{R}^{C \times 1 \times 1}$. Specifically, $M_{c}$ is calculated as follows
\begin{equation}\label{eq_2}
     M_{c} = S_n(AvgPool(F)) + S_n(MaxPool(F))
\end{equation}
In addition, the MT gate is equivalent to a convolution layer with fixed kernel coefficients. We highlight the manipulation traces in the feature maps via the MT gate to obtain the attention map $M_{t}$. Specifically, $M_{t}$ is calculated as
\begin{equation}\label{eq_3}
     M_{t}=\sum_{j=1}^{N}F_{j} \ast G
\end{equation}
Finally, we aggregate two attention maps, namely $M_{c}$ and $M_{t}$, to obtain the refined feature map $F'$.
\begin{equation}\label{eq_3}
     F'=\sigma(M_{c}) + \sigma(\alpha \cdot M_{t})
\end{equation}
where $\sigma$ represents the sigmoid function, and $\alpha$ denotes the coefficient that can be adaptively updated during back propagation.

\subsection{Network Architecture}
For the architecture of the proposed GocNet, both TP and MTA modules are embedded into a dual-stream architecture, as shown in Fig. 3. Resnet-18 is used as the backbone network for each stream. Note that for the first stream, the TP module is used as pre-processing to refine the input feature tensor, and the refined feature tensor is fed into the backbone network to learn discriminant features. For the second stream, the MTA module is embedded into each basic block in the backbone network, which forces the network to focus on manipulation traces. Finally, the features of the two streams are fused by using the element-wise sum, which are sent to the full connection layer for classification.

\section{Experiments}
Subsection A describes the five benchmark datasets. Subsection B reports the experimental settings including evaluation metrics, baseline models, and implementation details. Subsection C shows the performance of the proposed module. Subsection D reports the ablation study of the GocNet. Subsection E compares the GocNet with some the state-of-the-art models. Finally, we also verify the generalization ability of the GocNet in the Subsection F.

\begin{figure*}
  \centering
  \includegraphics[width=6.2in]{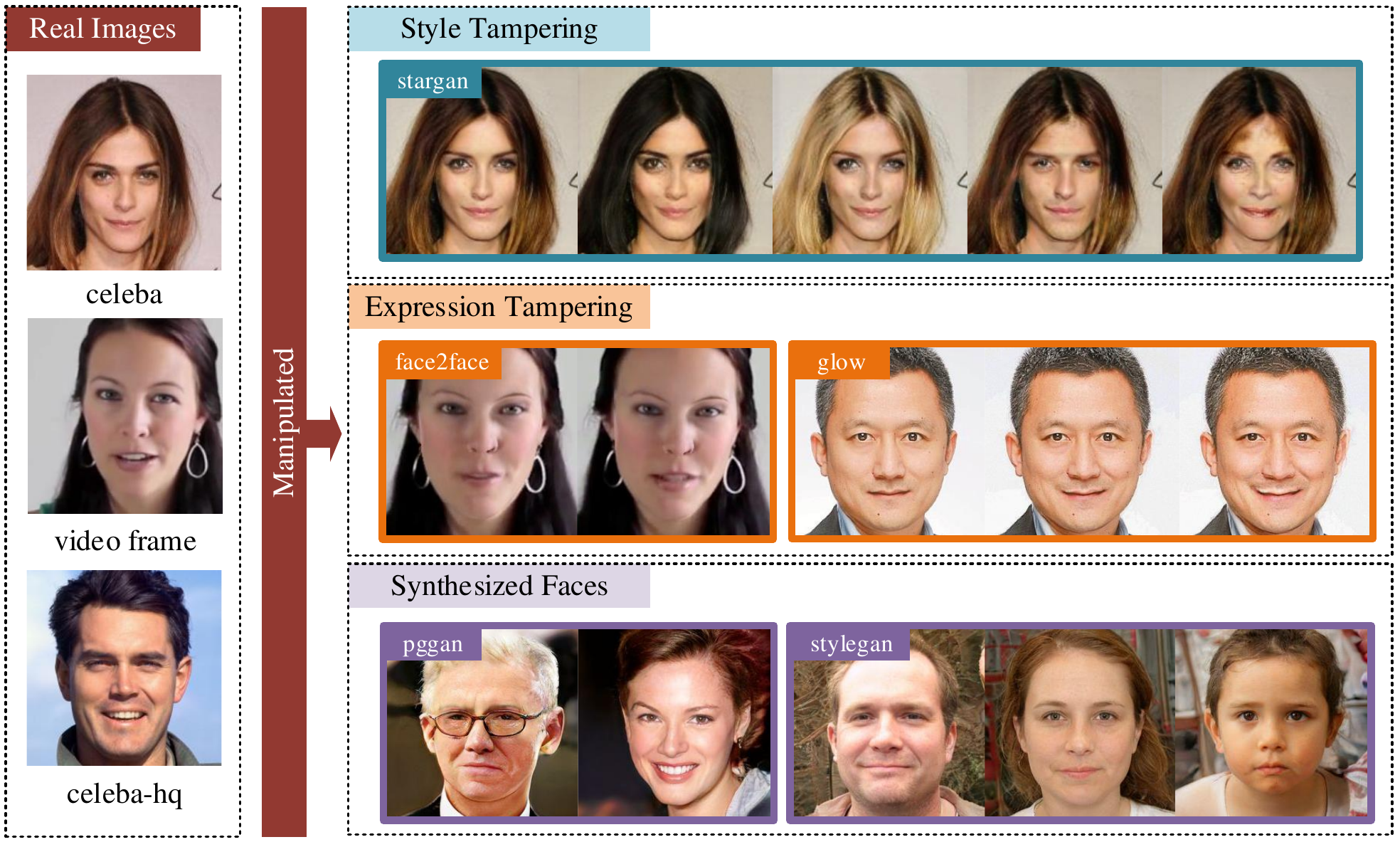}
  \caption{Some face images from the HFF dataset.}\label{HFF_dataset}
\end{figure*}

\subsection{Benchmark Dataset}
The proposed GocNet is evaluated on five datasets including FaceForensics++ (FF++) \cite{FaceForensics++}, Deep Fake Detection (DFD) \cite{DFD}, DeepFake Detection Challenge (DFDC) \cite{DFDC}, Hybrid Fake Face (HFF) \cite{AMTEN}, and Celeb-DF (CDF) \cite{CelebDF}.

\textbf{The FF++ dataset} is one of the most popular face forensics datasets, which contains 1k original video sequences and 4k fake video sequences manipulated by four methods including DeepFakes, Face2Face, FaceSwap and NeuralTextures. This dataset provides video sequences with three quality levels, which are raw, high-quality (HQ) and low-quality (LQ), respectively. To facilitate the experiments, we only use visually lossless HQ videos and visually lossy LQ videos, which are widely used in social networks. Both HQ and LQ videos are compressed by H.264/AVC, and the quantization parameters are set with 23 and 40, respectively.

\textbf{The DFD dataset} is a video dataset provided by Google \& JigSaw, which contains 3,068 deepfake video sequences. Note that they come from the real video sequences of the FF++ dataset. The DFD dataset also provides video sequences with three quality levels. In the experiments, we also use only visually lossless HQ videos and visually lossy LQ videos.

\textbf{The DFDC dataset} is a large-scale deepfake video dataset, which contains more than 100k real and fake video sequences (about 470GB). Note that real video sequences are manipulated by two Deepfake algorithms to obtain fake video sequences. We randomly select 2,891 real videos and 20,210 fake videos for our experiments.

\textbf{The HFF dataset} is another large-scale AI-generated image dataset, which contains 155k face images with high visual qualities. In this dataset, the real images come from the CelebA dataset \cite{celeba}, the CelebA-HQ dataset \cite{pggan}, and the YouTube video frames \cite{Faceforensics}, respectively. The fake images are generated by Glow \cite{glow}, Face2Face\cite{Face2Face}, StarGAN \cite{StarGAN}, PGGAN \cite{pggan}, and StyleGAN \cite{stylegan}, respectively. Among them, Glow and Face2Face are used for facial expression manipulation, StarGAN are used for changing facial attributives and styles such as hair color and gender, whereas PGGAN and StyleGAN are used for synthesizing face images with the spatial resolution up to 1024$\times$1024.

\textbf{The CDF dataset} is the most recent yet challenging large-scale deepfake video dataset. It contains 5,639 deepfake video sequences with high visual qualities. Note that the video sequences are generated by the improved deepfake methods. Thus, the CDF dataset has greatly reduced the gap in visual quality with those deepfake videos circulating over the Internet. In our experiments, we expoit this dataset to test the generalization ability of the proposed GocNet.

\begin{figure*}
  \centering
  \includegraphics[width=6.2in]{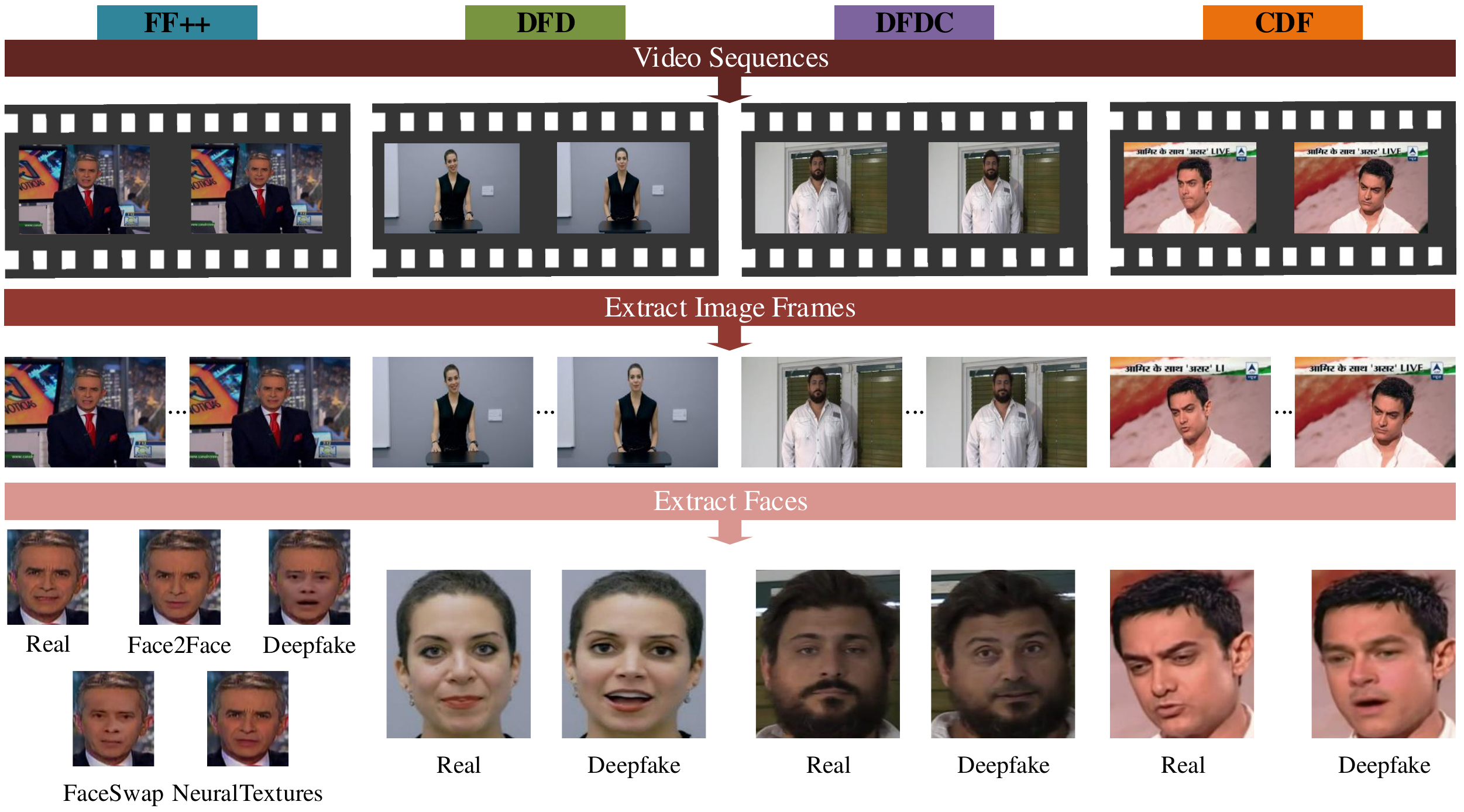}
  \caption{Sample images from four video datasets. }\label{video_datasets}
\end{figure*}

Among these five datasets, HFF is an image dataset, while the other four are video datasets. For the HFF dataset, the sizes of all images are resized to 299$\times$299. For the video datasets, we extract some face frames from each dataset and resize their sizes into 299$\times$299 by image cropping for the experiments.
Fig. \ref{HFF_dataset} and Fig. \ref{video_datasets} show some sample images of the HFF dataset and the other four video datasets, respectively. Table \ref{tbl:datasets} summarizes the details of each dataset.

\begin{table}[]
\centering
\caption{Five benchmark datasets for experiments.}
\label{tbl:datasets}
\resizebox{85mm}{!}{
\begin{tabular}{@{}c|c|c|cc|c@{}}
\toprule[1.2pt]
Datasets              & Classification        & Manipulate method & Training data    & Testing data & Total                \\ \midrule \midrule
\multirow{5}{*}{FF++} & Real                  & \textbackslash{}  & 50k              & 10k          & 60k                  \\
                      & \multirow{4}{*}{Fake} & Face2Face         & 12.5k            & 2.5k         & \multirow{4}{*}{60k} \\
                      &                       & FaceSwap          & 12.5k            & 2.5k         &                      \\
                      &                       & DeepFake          & 12.5k            & 2.5k         &                      \\
                      &                       & NeuralTextures    & 12.5k            & 2.5k         &                      \\ \hline \hline
\multirow{2}{*}{DFD}  & Real                  & \textbackslash{}  & 50k              & 10k          & 60k                  \\
                      & Fake                  & DeepFake          & 50k              & 10k          & 60k                  \\ \hline \hline
\multirow{2}{*}{DFDC} & Real                  & \textbackslash{}  & 50k              & 10k          & 60k                  \\
                      & Fake                  & DeepFake          & 50k              & 10k          & 60k                  \\ \hline \hline
\multirow{6}{*}{HFF}  & Real                  & \textbackslash{}  & 48k              & 12k          & 60k                  \\
                      & \multirow{5}{*}{Fake} & Face2Face         & 20k              & 5k           & \multirow{5}{*}{95k} \\
                      &                       & Glow              & 20k              & 5k           &                      \\
                      &                       & PGGAN             & 8k               & 2k           &                      \\
                      &                       & StarGAN           & 20k              & 5k           &                      \\
                      &                       & StyleGAN          & 8k               & 2k           &                      \\ \hline \hline
\multirow{2}{*}{CDF}                   & Real                  & \textbackslash{}  & \textbackslash{} & 60k          & 60k                  \\
                      & Fake                  & DeepFake          & \textbackslash{} & 60k          & 60k                  \\ \toprule[1.2pt]
\end{tabular}
}
\end{table}

\subsection{Experimental Settings}
\subsubsection{Evaluation Criterion}
For performance evaluation, there are three widely-used metrics, namely accuracy rate (ACC), the area under ROC curve (AUC), and equal error rate (EER). In the experiments, they are used to evaluate all the models, including the proposed GocNet and baseline models. Note that EER is achieved when the false rejection rate (FRR) is equal to the false acceptance rate (FAR). Thus, the smaller the EER value, the higher the overall performance of the model.

\subsubsection{Baseline Models}
We select several existing works as the baseline models for comparisons, which are summarized as follows.
\begin{itemize}
  \item Meso-4 \cite{MesoNet}: It exploits the mesoscopic properties of face images to expose facial forgeries.
  \item MesoInception-4 \cite{MesoNet}: It improves the Meso-4 by introducing the inception modules, achieving better performance for face forensics tasks.
  \item HP-CNN \cite{HPcnn}: Three high-pass filters are used as the pre-processing to highlight manipulation traces. And the processed images are input into a designed CNN model. In the experiments, the high-pass filter with the best performance is used for comparisons.
  \item MISLnet \cite{constrainedCNN}: It designs a C-Layer, which is combined with CNN for general-purpose image forensics.
  \item XceptionNet \cite{Xception}: It has been used face forgery detection and achieved desirable performance.
  \item AMTENnet \cite{AMTEN}: It designs an AMTEN and combines it with CNN to expose facial forgeries.
  \item BaseNet \cite{ResNet}: To prove the performance gains brought by two modules, we use the Resnet-18 as BaseNet.
  \item AMTEN-BaseNet \cite{AMTEN}: It refers to the combination of BaseNet with AMTEN that serves as the pre-processing module.
  \item C-Layer-BaseNet \cite{constrainedCNN}: It refers to the combination of BaseNet with the C-Layer as the pre-processing.
  \item TP-BaseNet: It refers to the comination of BaseNet with the TP module, in which nine gradient operators are used for comparisons.
  \item BaseNet-CBAM \cite{CBAM}: It refers to BaseNet which uses the convolutional block attention module (CBAM) for feature refinement.
  \item BaseNet-DA \cite{DAnet}: It refers to BaseNet which exploits a dual attention (DA) module for feature refinement.
  \item BaseNet-MTA: It refers to BaseNet which exploits our MTA module for feature refinement.
  \item BaseNet-MTA-Conv: It refers to the combination of BaseNet with the MTA module, yet the MT gate in the MTA module is replaced with plain convolution layer.
\end{itemize}

\subsubsection{Implementation Details}
We use two NVIDIA GeForce GTX 1080 Ti GPU to train GocNet under PyTorch framework. The GocNet employs the Adam with the parameters ($\beta_1 = 0.9$, $\beta_2 = 0.999$) for network optimization \cite{Adam}. In addition, an exponential learning rate ($\gamma = 0.5$) with an initial learning rate of 0.0005 is used to train the model. During the training process, some data augmentation operations such as rotation, normalization, horizontal flip, and random perspective are used to improve the model robustness.

\subsection{Key Module Comparison}
In this subsection, we compare the proposed modules with existing modules. The LQ video sequences in the FF++ dataset are selected for experiments.

\subsubsection{Comparison of the Pre-processing Modules} For the CNN-based image forensics, it is a common practice to use pre-processing to highlight manipulation traces and avoid the side effects due to distinct image content. For fair comparisons, different pre-processing modules are combined with the same backbone network (Resnet-18), respectively. Specifically, we compare the TP module with the state-of-the-art pre-processing works such as AMTEN \cite{AMTEN} and C-Layer \cite{constrainedCNN} in our experiments. In addition, nine classical gradient operators are used in the TP module to compare their performance, respectively.

\begin{table*}[]
\centering
\caption{Comparisons of the pre-processing modules.}
\label{tbl:pre}
\resizebox{160mm}{!}{
\begin{tabular}{@{}c|c|ccc@{}}
\toprule[1.2pt]
Models          & Description                                      & ACC(\%)        & AUC(\%)        & EER             \\ \midrule \midrule
BaseNet \cite{ResNet}         & backbone network                                 & 82.69          & 85.80          & 0.2326          \\ \hline \hline
AMTEN-BaseNet \cite{AMTEN}   & \multirow{2}{*}{existing pre-processing modules} & 83.41          & 87.01          & 0.2225          \\
C-Layer-BaseNet \cite{constrainedCNN} &                                                  & 83.61          & 88.77          & 0.1960          \\ \hline \hline
TP-BaseNet-1    & using highpass operator                          & 82.83          & 87.77          & 0.2037          \\
TP-BaseNet-2    & using robert sharpening operator                 & 86.20          & 91.14          & 0.1705          \\
TP-BaseNet-3    & using krisch operator                            & 87.62          & 92.73          & 0.1539          \\
TP-BaseNet-4    & using laplacian operator                         & 82.58          & 87.07          & 0.2129          \\
TP-BaseNet-5    & using sobel operator in horizontal direction     & 85.45          & 89.53          & 0.1884          \\
TP-BaseNet-6    & using sobel operator in vertical direction       & 87.36          & 92.32          & 0.1592          \\
TP-BaseNet-7    & using prewitt operator in horizontal direction   & 86.43          & 91.13          & 0.1673          \\
TP-BaseNet-8    & using prewitt operator in vertical direction     & 87.91          & 92.45          & 0.1550          \\
TP-BaseNet-9    & using prewitt operator in diagonal direction     & \textbf{88.21} & \textbf{93.00} & \textbf{0.1480} \\ \toprule[1.2pt]
\end{tabular}
}
\end{table*}

Table \ref{tbl:pre} reports the experimental results when using different pre-processing modules. From it, we can observe that different pre-processing modules bring different performance gains. On the basis of BaseNet, AMTEN brings about 0.72\% and 1.21\% performance gains in terms of ACC and AUC, and reduces 0.0101 in EER. Compared with AMTEN, the C-Layer module improves more detection performance. In essence, the existing pre-processing modules suppresses image content by complex learning mechanisms, so that the backbone network can learn features from manipulation traces instead of image content. Nevertheless, the complex learning mechanism might destroy the original manipulation traces when suppressing image content, which limits the detection performance. Thus, we exploit the gradient operators in the TP module to refine the feature tensor, which directly highlights subtle manipulation traces. Table \ref{tbl:pre} also compares the nine gradient operators when they are used in TP module.
From it, the prewitt operator in diagonal direction achieves the best detection results when it is used as the pre-processing.
Compared with BaseNet, TP-BaseNet brings performance gains about 5.52\% and 7.20\% in terms of ACC and AUC, and reduces 0.0846 in EER. From the experimental results, we conclude that the gradient operator does refine the feature tensor, and outperforms the existing pre-processing modules.
Since the prewitt operator in diagonal direction achieved the best results, it is used for the TP module in subsequent experimental comparisons.

\subsubsection{Comparison of the Attention Modules}
For the CNN-based computer vision and image forensics tasks, defining an appropriate attention mechanism is a common and effective way to improve classification accuracy. The proposed MTA module is compared with the existing attention modules such as CBAM \cite{CBAM} and DA \cite{DAnet}. Specifically, three attention modules are embedded into the same backbone network, respectively. In addition, to further prove the effectiveness of the MT gate in the MTA module, it is replaced by plain convolution layer for performance comparisons.

Table \ref{tbl:att} compares the detection results when using three attention modules. From it, we can observe that three
attention modules bring different performance gains to BaseNet. Since both CBAM and DA module are designed to improve feature representation for computer vision tasks such as image classification, they pay more attention to the pixel interdependencies of space and channel dimensions, instead of the manipulation traces. This constrains their performance gains for image forensic tasks. For the MTA module, the MT gate is specially designed to focus on the manipulation traces. Thus, BaseNet-MTA achieves the best accuracies, which proves that the MTA module is superior to existing attention modules for image forensics. Moreover, from the last two rows of Table \ref{tbl:att}, the effectiveness of the MT gate is verified for the MTA module.

\begin{table}
\centering
\caption{Comparisons of three attention modules.}
\label{tbl:att}
\resizebox{85mm}{!}{
\begin{tabular}{@{}c|ccc@{}}
\toprule[1.2pt]
Models           & ACC(\%)        & AUC(\%)        & EER             \\ \midrule \midrule
BaseNet \cite{ResNet}          & 82.69          & 85.80          & 0.2326          \\
BaseNet-CBAM \cite{CBAM}     & 83.62          & 88.27          & 0.2110          \\
BaseNet-DA \cite{DAnet}       & 85.78          & 89.86          & 0.1931          \\ \hline \hline
BaseNet-MTA-conv & 83.78          & 88.43          & 0.2086          \\
BaseNet-MTA      & \textbf{86.61} & \textbf{91.47} & \textbf{0.1761} \\ \toprule[1.2pt]
\end{tabular}
}
\end{table}

\subsection{Ablation Study}
In this work, both TP and MTA modules are two innovations for the proposed GocNet. In the network, two modules are used in different positions. Nevertheless, repeated refinement of features might have side effects on classification performance.
To combine two modules more effectively, we conduct ablation experiments on single-stream and dual-stream architectures. Note that each stream adopts Resnet-18 as the backbone network, and the dataset used in the experiment is the same with Subsection 4.3.

\begin{table}[]
\centering
\caption{Ablation study of the TP and MTA modules on single-stream architecture.}
\label{tbl:s-ablation}
\resizebox{87mm}{!}{
\begin{tabular}{@{}c|cc|ccc@{}}
\toprule[1.2pt]
Architecture                   & TP module        & MTA module       & ACC(\%)        & AUC(\%)        & EER             \\ \midrule \midrule
\multirow{4}{*}{Single-stream} & \textbackslash{} & \textbackslash{} & 82.69          & 85.80          & 0.2326          \\
                               & $\checkmark$     & \textbackslash{} & \textbf{88.21} & \textbf{93.00} & \textbf{0.1480} \\
                               & \textbackslash{} & $\checkmark$     & 86.61          & 91.47          & 0.1761          \\
                               & $\checkmark$     & $\checkmark$     & 83.90          & 87.32          & 0.2104          \\ \toprule[1.2pt]
\end{tabular}
}
\end{table}

\begin{table}[]
\centering
\caption{Ablation study of the TP and MTA modules on dual-stream architecture.}
\label{tbl:d-ablation}
\resizebox{87mm}{!}{
\begin{tabular}{@{}c|cc|ccc@{}}
\toprule[1.2pt]
Architecture                 & \begin{tabular}[c]{@{}c@{}}TP module\\ (first stream)\end{tabular} & \begin{tabular}[c]{@{}c@{}}MTA module\\ (second stream)\end{tabular} & ACC(\%)        & AUC(\%)        & EER             \\ \midrule \midrule
\multirow{4}{*}{Dual-stream} & \textbackslash{}                                                   & \textbackslash{}                                                     & 85.14          & 87.86          & 0.2154          \\
                             & $\checkmark$                                                       & \textbackslash{}                                                     & 87.75          & 91.09          & 0.1735          \\
                             & \textbackslash{}                                                   & $\checkmark$                                                          & 86.58          & 90.16          & 0.1889          \\
                             & $\checkmark$                                                       & $\checkmark$                                                         & \textbf{89.59} & \textbf{94.21} & \textbf{0.1379} \\ \toprule[1.2pt]
\end{tabular}
}
\end{table}

Table \ref{tbl:s-ablation} reports the ablation experiments on single-stream architecture. From it, we can observe that when either TP or MTA is embedded into the network, desirable detection results are achieved. However, when two modules are simultaneously embedded in single-stream architecture, there are performance degradations. The reasons behind this are summarized as follows. In the pre-processing stage, the TP module suppresses image content and highlights manipulation traces. However, when the pre-processed tensor is fed into the MTA module, it might interfere with the manipulation traces, leading to performance loss. The pre-processed tensor can be regarded as a tensor containing less information. Apparently, it is better for the MTA module to extract features from tensor with richer information.

To avoid the interaction between the TP module and the MTA module, further experiments are done on the dual-stream architecture. The first stream only receives the refined feature tensor from the TP module, and the second stream embeds the MTA module to learn manipulation trace features from RGB images. Table \ref{tbl:d-ablation} reports the experimental results on the dual-stream architecture. From it, we can observe that using either TP or MTA module in the dual-stream network improves the performance.
However, when both TP and MTA modules are used in the network, the best performance is achieved, in which the ACC is 89.59\%, the AUC is 94.21\%, and the EER is 0.1379. Thus, we adopt the dual-stream GocNet equipped with both TP and MTA for subsequent experimental comparisons.

\subsection{Comparison with Prior Works}
To further prove the effectiveness of the proposed GocNet, it is compared with the existing works on four datasets including FF++, DFD, DFDC, and HFF. Table \ref{tbl:FF++} to Table \ref{tbl:HFF} report the results of binary classification (Real or Fake) on these four datasets, respectively.

\begin{table}[]
\centering
\caption{Performance evaluation for different forensics models on the FF++ dataset.}
\label{tbl:FF++}
\resizebox{87mm}{!}{
\begin{tabular}{@{}c|cccccc@{}}
\toprule[1.2pt]
\multirow{2}{*}{Models}      & \multicolumn{2}{c}{ACC(\%)}     & \multicolumn{2}{c}{AUC(\%)}     & \multicolumn{2}{c}{EER}           \\
            & C23            & C40            & C23            & C40            & C23             & C40             \\ \midrule \midrule
Meso-4 \cite{MesoNet}     & 54.28          & 53.82          & 55.82          & 54.27          & 0.4591          & 0.4727          \\
Meso-Incep \cite{MesoNet} & 79.52          & 76.37          & 79.69          & 78.55          & 0.2773          & 0.3099          \\
HP-CNN \cite{HPcnn}     & 74.81          & 71.43          & 80.97          & 76.03          & 0.2761          & 0.3085          \\
XceptionNet \cite{FaceForensics++} & 88.56          & 82.80          & 91.38          & 85.97          & 0.1780          & 0.2322          \\
AMTENnet \cite{AMTEN}   & 82.69          & 78.76          & 83.47          & 80.93          & 0.2567          & 0.2754          \\
MISLnet \cite{constrainedCNN}    & 90.58          & 83.78          & 87.31          & 83.82          & 0.2184          & 0.2415          \\ \hline \hline
GocNet      & \textbf{94.86} & \textbf{89.59} & \textbf{97.55} & \textbf{94.21} & \textbf{0.0855} & \textbf{0.1379} \\ \toprule[1.2pt]
\end{tabular}
}
\end{table}

\begin{table}[]
\centering
\caption{Performance evaluation for different forensics models on the DFD dataset.}
\label{tbl:DFD}
\resizebox{87mm}{!}{
\begin{tabular}{@{}c|cccccc@{}}
\toprule[1.2pt]
\multirow{2}{*}{Models}      & \multicolumn{2}{c}{ACC(\%)}     & \multicolumn{2}{c}{AUC(\%)}     & \multicolumn{2}{c}{EER}           \\
            & C23            & C40            & C23            & C40            & C23             & C40             \\ \midrule \midrule
Meso-4 \cite{MesoNet}     & 63.33          & 70.43          & 51.71          & 57.08          & 0.4930          & 0.4657          \\
Meso-Incep \cite{MesoNet} & 87.85         & 89.47          & 96.23          & 94.78          & 0.1029          & 0.1230          \\
HP-CNN \cite{HPcnn}     & 92.55          & 89.71          & 96.53          & 92.47          & 0.0956          & 0.1518          \\
XceptionNet \cite{FaceForensics++} & 96.57          & 94.84          & 98.92          & 97.82          & 0.0535          & 0.0763          \\
AMTENnet \cite{AMTEN}   & 95.69          & 93.29          & 97.94          & 95.93          & 0.0715          & 0.1080          \\
MISLnet \cite{constrainedCNN}    & 96.11          & 93.16          & 98.24          & 96.05          & 0.0730         & 0.1041          \\ \hline \hline
GocNet      & \textbf{97.77} & \textbf{96.06} & \textbf{99.46} & \textbf{98.55} & \textbf{0.0363} & \textbf{0.0622} \\ \toprule[1.2pt]
\end{tabular}
}
\end{table}

\begin{table}[]
\centering
\caption{Performance evaluation for different forensics models on the DFDC dataset.}
\label{tbl:DFDC}
\resizebox{80mm}{!}{
\begin{tabular}{@{}c|ccc@{}}
\toprule[1.2pt]
Models               & ACC(\%)        & AUC(\%)        & EER             \\ \midrule \midrule
Meso-4 \cite{MesoNet} & 60.91          & 62.53          & 0.4125          \\
Meso-Incep \cite{MesoNet} & 80.00          & 84.34          & 0.2379          \\
HP-CNN \cite{HPcnn}     & 70.87          & 77.68          & 0.2949          \\
XceptionNet \cite{FaceForensics++} & 90.05          & 93.22          & 0.1510          \\
AMTENnet \cite{AMTEN}   & 86.67          & 87.05          & 0.2135          \\
MISLnet \cite{constrainedCNN}    & 90.43          & 92.29          & 0.1583          \\ \hline \hline
GocNet      & \textbf{92.57} & \textbf{95.97} & \textbf{0.1107} \\ \toprule[1.2pt]
\end{tabular}
}
\end{table}

For face image forensics, improving feature representation is the key issue to enhance detection performance. Generally, there are two ways to enhance feature representation. First, constructing a complex convolution network with powerful feature representation by reasonably stacking convolution layers. Second, enforcing the network to learn more discriminative features directly from manipulation traces by embedding specific modules into the network. In our experiment, we select six representative methods for experimental comparisons. Among them, the first category of approaches include Meso-4, Meso-Incep, and XceptionNet, and the second category of approaches include HP-CNN, AMTENnet, and MISLnet.

From the results reported in Table \ref{tbl:FF++} to Table \ref{tbl:HFF}, we can observe that embedding specific modules into the network usually achieves better detection results than stacking convolution layers. Note that Meso-4 and Meso-Incep are shallow CNNs, so they have relatively low accuracies in four groups of experiments. XceptionNet, which is a deep CNN with complex structure, achieves competitive results. However, the complex network structure usually needs a lot of training time. Thus, it is not an ideal way to find the optimal solution of feature representation by constructing complex network structures.
Moreover, HP-CNN, AMTENnet, and MISLnet use different pre-processing techniques to highlight the manipulation traces. Though these pre-processing techniques are only combined with shallow CNNs, promising results have been achieved. Especially, benefiting from the TP and MTA modules, GocNet achieves the best detection results on four datasets. Note that though AMTEN and C-Layer are two premium tools for highlighting manipulation traces, TP and MTA are more advanced tools for this purpose.

\subsection{Generalization Ability}
To further validate the generalization capability, the proposed GocNet is tested on a latest benchmark dataset. That is, GocNet is trained on the FF++ dataset with two compression levels as a binary detector, which are called GocNet-HQ and GocNet-LQ, respectively. Then, the pre-trained GocNet-HQ and GocNet-LQ are tested on the CDF dataset, respectively.
Table \ref{tbl:CDF} reports the AUC scores of various detection methods. Apparently, GocNet achieves promising results, especially the AUC score of GocNet-LQ trained with low-quality data has reached 67.43\%.

\begin{table}[]
\centering
\caption{Performance evaluation for different forensics models on the HFF dataset.}
\label{tbl:HFF}
\resizebox{80mm}{!}{
\begin{tabular}{@{}c|ccc@{}}
\toprule[1.2pt]
Models               & ACC(\%)        & AUC(\%)        & EER             \\ \midrule \midrule
Meso-4 \cite{MesoNet} & 61.04          & 52.74          & 0.4775          \\
Meso-Incep \cite{MesoNet} & 98.61          & 99.45          & 0.0374          \\
HP-CNN \cite{HPcnn}     & 96.33          & 99.12          & 0.0504          \\
XceptionNet \cite{FaceForensics++} & 99.06          & 99.68          & 0.0303          \\
AMTENnet \cite{AMTEN}   & 99.82          & 99.87          & 0.0160          \\
MISLnet \cite{constrainedCNN}    & 98.40          & 99.15          & 0.0519          \\ \hline \hline
GocNet      & \textbf{99.87} & \textbf{99.95} & \textbf{0.0111} \\ \toprule[1.2pt]
\end{tabular}
}
\end{table}

\section{Conclusion}
In this work, we rethink the classical gradient operators for exposing AI-generated face images. Two modules, namely TP and MTA, are designed to highlight the manipulation traces left by face forgeries. Moreover, a dual-stream network model that equipped with the TP and MTA modules, namely GocNet, is proposed for exposing AI-generated face images.
We systematically validate the performance of the proposed approaches on five open datasets. The extensive experimental results show that GocNet achieves better results than the existing works. Moreover, the experiments prove that TP and MTA are advanced tools for highlighting manipulation traces. Especially, the TP module greatly improves the detection performance by simply convolving the feature tensor with the gradient operator, which avoids the loss caused by the complex prediction mechanism in the existing works.

\begin{table}[]
\centering
\caption{The AUC scores(\%) of various detection methods on Celeb-DF dataset.}
\label{tbl:CDF}
\resizebox{87mm}{!}{
\begin{tabular}{@{}c|cc@{}}
\toprule[1.2pt]
Methods         & Training dataset        & Celeb-DF \cite{CelebDF}      \\ \midrule \midrule
Two-stream \cite{zhoupeng}     & Private dataset         & 53.8           \\
Meso-4 \cite{MesoNet}         & Private dataset         & 54.8           \\
MesoInception-4 \cite{MesoNet} & Private dataset         & 53.6           \\
HeadPose \cite{head_poses}       & UADFV dataset           & 54.6           \\
FWA \cite{SimulateArtifacts}            & UADFV dataset           & 56.9           \\
VA-MLP \cite{visual_artifacts}         & Private dataset         & 55.0           \\
VA-LogReg \cite{visual_artifacts}      & Private dataset         & 55.1           \\
Xception-raw \cite{FaceForensics++}   & FaceForensics++ dataset & 48.2           \\
Xception-HQ \cite{FaceForensics++}    & FaceForensics++ dataset & 65.3           \\
Xception-LQ \cite{FaceForensics++}    & FaceForensics++ dataset & 65.5           \\
Multi-task \cite{nguyen2019multi}     & FaceForensics dataset   & 54.3           \\
Capsule \cite{Capsule2}        & Private dataset         & 57.5           \\ \hline \hline
GocNet-HQ       & FaceForensics++ dataset & 59.46          \\
GocNet-LQ       & FaceForensics++ dataset & \textbf{67.43} \\ \toprule[1.2pt]
\end{tabular}
}
\end{table}

\ifCLASSOPTIONcaptionsoff
  \newpage
\fi


\bibliography{refs}
\bibliographystyle{IEEEtrans}

\end{document}